\documentclass{article}
\usepackage{lmodern}

\usepackage[final]{colm2025_conference}
\usepackage{silence}
\usepackage{float}
\usepackage{microtype}
\usepackage{hyperref}
\usepackage{url}
\usepackage{booktabs}
\usepackage{lineno}
\usepackage{graphicx}
\usepackage{silence}
\usepackage{amsmath,amsfonts,bm}
\usepackage[font=small,labelfont=bf]{caption}
\usepackage{amssymb} 
\usepackage{titlesec} 

\definecolor{darkblue}{rgb}{0, 0, 0.5}
\hypersetup{colorlinks=true, citecolor=darkblue, linkcolor=darkblue, urlcolor=darkblue}

\title{Ask WhAI: \\Probing Belief Formation in Role-Primed LLM Agents}

\author{Keith Moore\thanks{Corresponding author: \texttt{kem1@stanford.edu}},
Jun W. Kim,
David Lyu,
Jeffrey Heo,
Ehsan Adeli \\
Department of Biomedical Data Science, Stanford University \\
\texttt{\{kem1,junwkim,dlyu,jeffheo,eadeli\}@stanford.edu}
}

\begin{document}

\ifcolmsubmission
\linenumbers
\fi

\maketitle

\begin{abstract}
We present \texttt{Ask WhAI}, a systems-level framework for inspecting and perturbing belief states in multi-agent interactions. The framework records and replays agent interactions, supports out-of-band queries into each agent’s beliefs and rationale, and enables counterfactual evidence injection to test how belief structures respond to new information.

We apply the framework to a medical case simulator notable for its multi-agent shared memory (a time-stamped electronic medical record (\texttt{EMR}) and an oracle agent (the \texttt{LabAgent}) that holds ground truth lab results revealed only when explicitly queried.

We stress-test the system on a multi-specialty diagnostic journey for a child with an abrupt-onset neuropsychiatric presentation. Large language model agents, each primed with strong role-specific priors (“act like a neurologist”, “act like an infectious disease specialist”), write to a shared medical record and interact with a moderator across sequential or parallel encounters. Breakpoints at key diagnostic moments enable pre- and post-event belief queries, allowing us to distinguish entrenched priors from reasoning or evidence-integration effects.

The simulation reveals that agent beliefs often mirror real-world disciplinary stances (including over-reliance on canonical studies and resistance to counter-evidence) and that these beliefs can be traced and interrogated in ways not possible with human experts. By making such dynamics visible and testable, \texttt{Ask WhAI} offers a reproducible way to study belief formation and epistemic silos in multi-agent scientific reasoning.
\end{abstract}

\section{Introduction}

Large language models (LLMs) are increasingly used to emulate expert roles by priming them with instructions such as “Act like an experienced cardiologist” or “Act like a senior electrical engineer.” This role prompt (persona) brings with it a large set of priors: patterns of reasoning, terminology, and domain-specific assumptions learned from opaque training data. These priors can guide reasoning in useful ways, but they can also bias interpretation of new evidence, anchor decisions prematurely, or limit consideration of alternative explanations.

This work began with a simple interaction. We prompted ChatGPT-4o with “You are a pediatric neurologist.” Surprisingly, the LLM adopted both the tone of an expert and displayed the kind of diagnostic rigidity we've seen in clinical disputes. This raised the question: could a role-primed LLM act as an epistemic proxy and potentially reveal why it holds a belief and what information might change its “mind”?

To explore this, we built two components. The first is a \textbf{medical case simulation tool} that emulates patient encounters using LLM agents in the roles of patient, doctor, and other specialists, each operating with strong, role-specific priors. The second is \texttt{Ask WhAI}, a belief debugger that attaches to the simulation to record and replay interactions, inspect belief states before and after key “diagnostic moments,” and inject counterfactual evidence to test how reasoning changes.

To evaluate this approach, we use a deliberately difficult test case: the multi-specialty evaluation of a child with a \textbf{rare, abrupt-onset neuropsychiatric presentation}.
This scenario is well-suited as a stress test because different specialties often interpret the same patient record in sharply divergent ways, even when shown identical labs and history. The combination of entrenched priors, sensitivity to context, and resistance to belief revision makes it an ideal proving ground for multi-agent belief inspection.

\textbf{Our goal} is not to resolve debate about any particular disorder 
(though it would be valuable if this work eventually contributed) but rather to test whether using LLMs as role-based agents can expose implicit assumptions captured by the models that might inform new research toward clarification. 
To that end, we employ a composite, literature-informed simulation of an abrupt-onset neuropsychiatric presentation to examine how encounter order and disciplinary priors shape diagnostic reasoning. 
The scenario \textbf{is illustrative} rather than clinical; all elements are fully synthetic and not drawn from or affiliated with any real patient, program, or institution. 
The simulator should be interpreted as an analytical probe of reasoning processes rather than a reproduction of clinical decision-making 
(see Appendix~\ref{appendix:agentic_arch} for system details and the Ethics Statement for scope and compliance).

\section{Related Work}

\textbf{Beyond accuracy:} Most evaluations of agentic systems focus on accuracy as the primary endpoint. In contrast, our goal is to inspect whether an agent can surface and articulate the beliefs guiding its reasoning, and whether those beliefs can be perturbed. This is closer to work in explainability and interpretability \citep{zhao2024explainability}, but with emphasis on role-specific priors: the assumptions and heuristics an LLM carries when primed as a particular expert. We seek to understand why the LLM holds a belief and if it can articulate that reason.

\textbf{Belief analysis tools:} Prior explainability work has largely examined single-agent reasoning through justification generation, output traceability, and post-hoc alignment checks. Related tools such as layered prompting \citep{faisal2025layeredmas}, AGDebugger \citep{epperson2025agdebugger}, and AgentRR \citep{feng2025agentrr} also support replay and message editing. We extend these by introducing a shared messaging framework that maintains a common interaction history (analogous to an EMR) across agents, while permitting out-of-band queries of internal belief state that do not interfere with the normal message flow. This enables contextualized inspection of belief formation within the same dialogue, rather than after-the-fact replay. A direct experimental comparison would conflate these distinct goals, but our approach can reproduce their replay functions while adding synchronized context sharing and introspective access absent in prior tools.

\textbf{Simulation frameworks:} Agentic medical simulators have introduced moderator roles \citep{swanson2024virtual}, persistent personas \citep{park2023generative}, structured encounters \citep{agrawal2023clinical}, and sequential diagnostic reasoning in the clinical setting \citep{nori2025sequentialdiagnosislanguagemodels}. Our system extends these by adding a shared, timestamped EMR for cross-agent memory, an oracle \texttt{LabAgent} for controlled evidence release, and a message cache for reproducible longitudinal simulations.

\section{System Overview}

\subsection{Medical Case Simulator}

\begin{figure}[ht]
  \centering
  \includegraphics[width=1.0\linewidth]{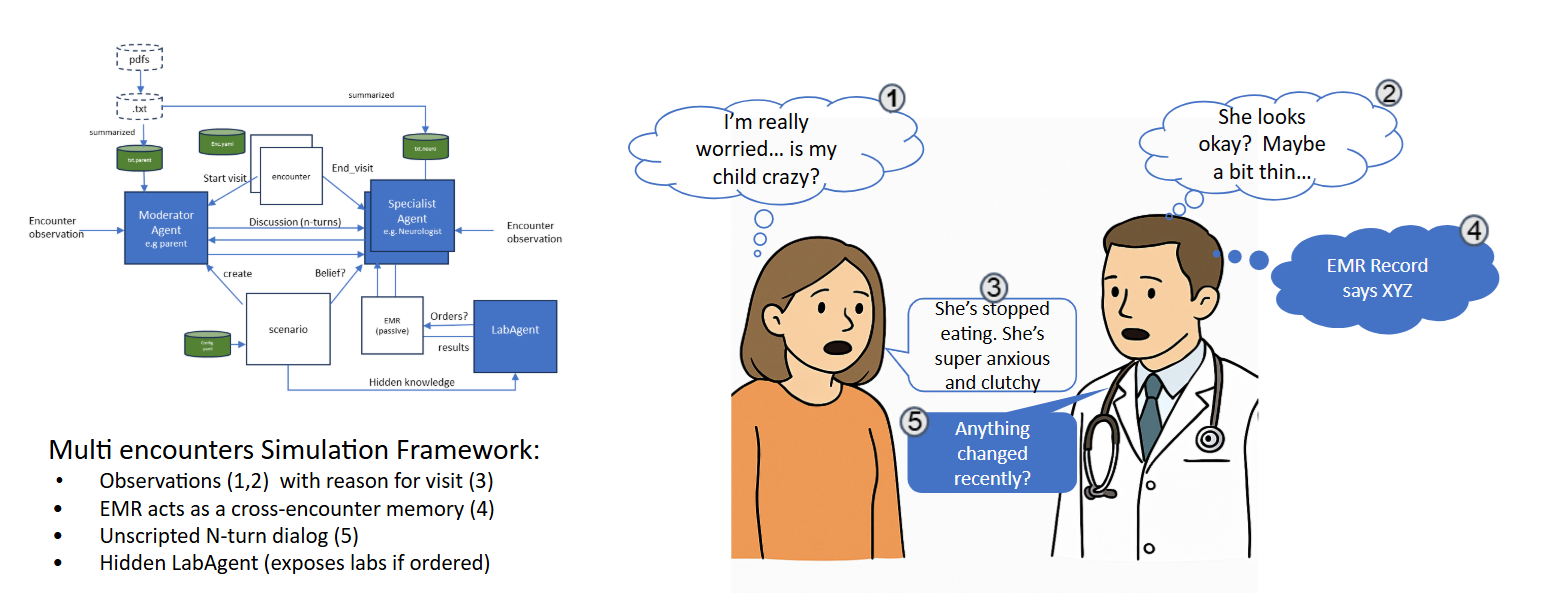}
  \caption{\small Conceptual architecture of an encounter between a moderator (e.g., parent) agent and a specialist agent.}
  \label{fig:encounter}
\end{figure}

The simulator models a diagnostic workflow as a sequence of encounters between large language model (LLM) agents assigned to distinct roles. Figure~\ref{fig:encounter} illustrates the architecture of a single encounter. Full system details, including the scripting format and overall architecture, are provided in Appendix~\ref{appendix:agentic_arch}.

Each encounter begins when the moderator agent\footnote{The architecture supports peer-to-peer messaging, but without a moderator, agents often talk past one another. Following \citet{swanson2024virtual}, we use a moderator (e.g., parent or lead reviewer) to keep discussions focused through challenges and rebuttals.} and the specialist internalize any private observations (steps 1 and 2). The specialist may also review the patient’s EMR (step 4), which serves as a timestamped, cross-agent memory. The moderator and specialist then engage in unscripted dialogue (step 5), triggered solely by the scripted \texttt{reason\_for\_visit} (step 3). All EMR updates and messages are logged.

The simulator supports:
\begin{itemize}
    \item Generating role-specific interactions from scripted scenarios
    \item Maintaining a longitudinal EMR with timestamped entries and evolving lab results
    \item Enforcing access controls based on role and encounter timing
\end{itemize}

\subsection{Ask WhAI Debugger}

\texttt{Ask WhAI} wraps the simulator to enable structured inspection and targeted perturbation of belief state. It treats each encounter boundary as a \emph{breakpoint} (i.e., a pause where agent beliefs can be queried, replayed, or modified). This enables us to trace belief evolution and test whether shifts are driven by evidence, framing, prompt design, or entrenched role-based priors.

Most experiments preserve the original sequence of encounters, varying only what information is shown, how it is framed, or how confidently it is expressed. For example, we may inject counter-belief documents, vary agent tone, or display lab results that were never originally ordered.

\texttt{Ask WhAI} also supports reordering encounters (i.e., visiting a psychiatrist before a neurologist); however, since these encounters surface test results, we generate lab tests when they are ordered. This means in some situations, a specialist is seeing a test result because a prior specialist ordered it.  This lets us probe how belief formation depends on content, attribution, \textbf{and timing}.

\texttt{Ask WhAI} exposes seven independent controls (Table~\ref{tab:controls}) that can be applied independently or jointly to simulate variation in workflow, tone, or evidence access. While we focus on encounter-level breakpoints, this approach can also apply to finer-grained events such as utterances or EMR edits.

\begin{table}[ht]
\small
\centering
\begin{tabular}{p{0.18\linewidth} p{0.7\linewidth}}
\toprule
\textbf{Control} & \textbf{Description} \\
\midrule
\textbf{Priming} & Provide agents with documents or evidence before an encounter. \\
\textbf{Exposing} & Reveal EMR state to agents before they engage. \\
\textbf{Probing} & Modify the belief query or the format of the response. \\
\textbf{Encounter order} & Reorder the sequence of specialist interactions. \\
\textbf{Lab results} & Insert or modify lab findings shown to the agent. \\
\textbf{Persona / voice} & Adjust the style or assertiveness of agent speech. \\
\textbf{Forced reflection} & Alter EMR prompts to elicit more reflective reasoning. \\
\bottomrule
\end{tabular}
\caption{Key controls available to influence specialist behavior and belief formation.}
\label{tab:controls}
\end{table}

At each breakpoint, Ask WhAI can:
\begin{itemize}
    \item Issue \textbf{pre-encounter queries} to assess prior beliefs or intended actions
    \item Issue \textbf{post-encounter queries} to capture belief updates and rationales
    \item \textbf{Replay} encounters deterministically or with modified context
    \item Inject \textbf{counterfactual evidence} to test alternate reasoning paths
    \item Compare belief states to isolate changes due to evidence vs. anchoring
\end{itemize}

\section{Use Case: Abrupt-Onset Neuropsychiatric Scenario}

\label{sec:usecase}

To demonstrate \texttt{Ask WhAI}, we simulate a multi-specialty diagnostic process for a child presenting with \textbf{handwriting changes}, \textbf{separation anxiety}, and \textbf{sudden-onset severe OCD} following a recent infection. This symptom constellation is consistent with a class of \textbf{abrupt-onset neuropsychiatric syndromes} known as \textbf{PANDAS} \citep{swedo1998pandas, Swedo2015} (Pediatric Autoimmune Neuropsychiatric Disorder Associated with Streptococcus). The simulated case serves as a useful stress test for our debugger because it centers on a long-standing controversy over causality (specifically, whether post-infectious mechanisms can trigger acute-onset psychiatric symptoms). This question has divided medical specialties for decades.\footnote{The association between infections and psychiatric symptoms has been noted since Osler’s study of obsessional behaviors following Sydenham’s chorea \citep{osler1894chorea}. Debate continues over whether such associations are causal or coincidental.}

Each specialist agent is initialized with a role-specific persona (e.g., ``you are a pediatric neurologist'') and receives identical access to the patient’s current EMR and lab results unless scenario controls specify otherwise. These role prompts encode strong domain-specific assumptions learned during training, which can anchor interpretation and bias evidence integration. The simulated case of abrupt onset symptoms offers an environment where such entrenched priors are likely to surface, allowing us to observe cases in which agents:
\begin{itemize}
    \item Ignore or discount relevant findings inconsistent with their priors
    \item Make logical errors even on evidence they accept
    \item Reach divergent conclusions despite having access to the same information
\end{itemize}

Table~\ref{tab:biases} summarizes the baseline diagnostic stance of each persona after the \emph{first} encounter, where all agents saw the same initial record and were asked whether an infection-triggered etiology was plausible.

\begin{table}[ht]
\centering
\small
\begin{tabular}{p{2.8cm} p{10.2cm}}
\toprule
\textbf{Agent Persona} & \textbf{Baseline perspective on PANDAS hypothesis (i.e., priors)} \\
\midrule
\textbf{Pediatric \newline Neurologist \newline Agent} & Infections are too common to infer causality, and abrupt symptom onset alone does not meet the bar for defining a new neurologic syndrome. \newline \textbf{Rationale:} Without identification of a specific autoantibody that modulates neuronal function or results from a well-controlled clinical trial, most neurologists remain skeptical. Exceptions like Sydenham chorea are accepted largely due to historical precedent, not emerging mechanisms. \\
\hline
\textbf{Pediatric \newline Rheumatologist \newline Agent} & Considers the case plausible based on analogy to other post-infectious inflammatory syndromes (e.g., Lyme arthritis, ARF, post-strep arthritis). \newline \textbf{Rationale:} Familiar with immune-mediated syndromes lacking a specific autoantibody. In rare pediatric conditions, treatment algorithms are often developed without randomized controlled trials. Acceptance is based on pattern recognition across cases. \\
\hline
\textbf{Child \newline Psychiatrist\newline Agent} & Frames the presentation as an abrupt but idiopathic onset of OCD or anxiety. Unless evidence emerges to justify a biologically distinct subtype, treatment proceeds using standard protocols. \newline \textbf{Rationale:} \citet{aacap2023ocd} guidelines for pediatric OCD recommend cognitive-behavioral therapy and SSRIs as first-line treatments. Absence of biomarkers or clear diagnostic separation keeps PANDAS within broader OCD management. \\
\hline
\textbf{General \newline Pediatrician \newline Agent} & May observe unusual presentation (e.g., sudden-onset OCD), but typically defers to subspecialists. \newline \textbf{Rationale:} Follows AAP guidelines \cite{AAP2025PANS} and refers out when presentation exceeds primary care scope. Limited exposure to post-infectious neuropsychiatric conditions during training. \\
\bottomrule
\end{tabular}
\caption{\small Summaries of perspective on PANDAS based on queries of agents before first encounter.}
\label{tab:biases}
\end{table}

These perspectives mirror real-world clinical encounters, where entrenched role-specific priors can lead to critical omissions. For example, if a child presents with contamination fears but no reported sore throat, the pediatrician may not consider strep as a possible cause and therefore not conduct a strep test. Similarly, when we asked the psychiatrist agent why a strep test had not been ordered, the reply was: “Because we don’t do strep tests. That is done by a pediatrician.” This highlights how divided responsibility (combined with implicit role boundaries) can prevent simple diagnostic actions, even when clinically relevant. 

This simulation is illustrative and not diagnostic, intended solely to test whether multi-agent systems reproduce characteristic reasoning dynamics observed in interdisciplinary settings.

\section{Experiments and Results}
\subsection{Reproducibility}
\label{sec:reproducibility}

Simulations are reproducible under fixed conditions by caching agent responses based on prompt content, model version (e.g., \texttt{gpt-4o-2024-08-06}), temperature, and message order. Timestamps and other non-behavioral elements are excluded from the cache key. Cached responses are reused unless bypassed explicitly (e.g., for stochastic probing). All cached outputs, scenario configurations, and simulator code are versioned and checked into our repository \citep{simrepo}.
\subsection{Experimental Setup}
Starting from the baseline beliefs in Table~\ref{tab:biases}, we ran a series of experiments, each varying a single debugger control (Table~\ref{tab:controls}).
Each experiment begins with an analytic question (“What do we want to vary?”), identifies a simulation \emph{hook}, modifies the simulator to expose it, and configures the debugger to isolate and measure its effect.

Belief is probed after each encounter using a configurable prompt.\footnote{Prompts are configured in a scenario \texttt{.yaml} file} In the first scenario, this was a simple categorical query (“Is this an infection-triggered etiology?”) with responses mapped to \{skeptical, neutral, believes\}. 

In the following subsection, we present three of the experiments (details of which are in the appendices).

\subsection{Experiment: Priming with Facts}
\label{sec:priming}
\textbf{Goal:} Pre-encounter document ingestion 

\textbf{Method:} Before an encounter, agents can be \textbf{primed} with documents (articles, papers, bulletins) that persist in memory. Because each agent summarizes through its persona, the same document can produce different internalizations. A special file, \texttt{0.txt}, is injected into the system prompt so all agents share baseline priming.

\textbf{Intervention:} Added to \texttt{0.txt} a recent bulletin stating that the American Academy of Pediatrics (AAP) recognized infection-triggered abrupt-onset neuropsychiatric syndrome as a research focus.

\textbf{Outcome:} The pediatrician moved from skeptical to \textbf{neutral} by encounter 3 and to \textbf{believes} by encounter 15, with all but the neurologist shifting stance (Table~\ref{tab:belief_evolution1}b).

\begin{table}[ht]
\centering
\small
\begin{tabular}{|c|c|c|c|c|}
\hline
\textbf{Encounter} & \textbf{a (Baseline)} &  \textbf{b Priming} & \textbf{c (Sherlock Mode)} & \textbf{d (Entangled)} \\
\hline
1 & Skeptical & Skeptical & Viral pharyngitis & Skeptical \\
\hline
3 & Skeptical & \textbf{Neutral} & PANS & Skeptical \\
\hline
15 & \textbf{Neutral} & \textbf{Believes} & \textbf{PANDAS} & \textbf{Believes} \\
\hline
\end{tabular}
\caption{\textbf{Pediatrician agent} belief evolution across identical scenarios.}
\label{tab:belief_evolution1}
\end{table}

\textbf{Observation:} Priming softened skepticism in most agents. Because the hook isolates document exposure, changes can be attributed to this intervention rather than uncontrolled history drift.

\subsection{Experiment: Information Exposure and Belief Assessment}
\label{sec:sherlock}
\textbf{Goal:} Out-of-band belief probing

\textbf{Method:} Using the same scenario as Section~\ref{sec:priming}, we compared three probe styles:
\begin{enumerate}
\item \textbf{Multiple Choice (a,b):} Ask if the case is infection triggered [skeptical, neutral, believes] and request justification.
\item \textbf{List-first (c):} Request top 10 diagnoses before any stance query.
\item \textbf{Entangled (d):} Ask for top diagnosis, then stance and justification for that diagnosis.
\end{enumerate}

\textbf{Intervention:} Changed only the probe format via the hook, leaving all other simulation elements fixed.

\textbf{Outcome:} In list-first probing, infection triggered surfaced earlier and stance shifted toward belief sooner. Entangled format produced early top-diagnosis commitment with explicit belief.

\textbf{Observation:} A single-word change (“diagnoses”) altered role behavior, overriding the cautious pediatrician persona. The hook made this effect measurable without polluting the EMR dialogue.

\subsection{Experiment: Order Effects on Belief Dynamics}
\label{sec:order}
\textbf{Goal:} Measure how belief formation in the pediatrician agent is shaped by the sequence and identity of prior specialist encounters.

\textbf{Method:} We ran six permutations of a fixed 16-encounter diagnostic scenario, varying the order of neurology, psychiatry, and rheumatology specialists. Pediatrician agents were inserted between specialist blocks and at the end to assess belief evolution. Belief scores (0–10) were derived from stance labels and averaged over three simulation runs at constant temperature.

\textbf{Intervention:} Only the specialist encounter order was varied; all other simulation parameters were held constant.

\begin{figure}[ht]
\centering
\includegraphics[width=0.7\linewidth]{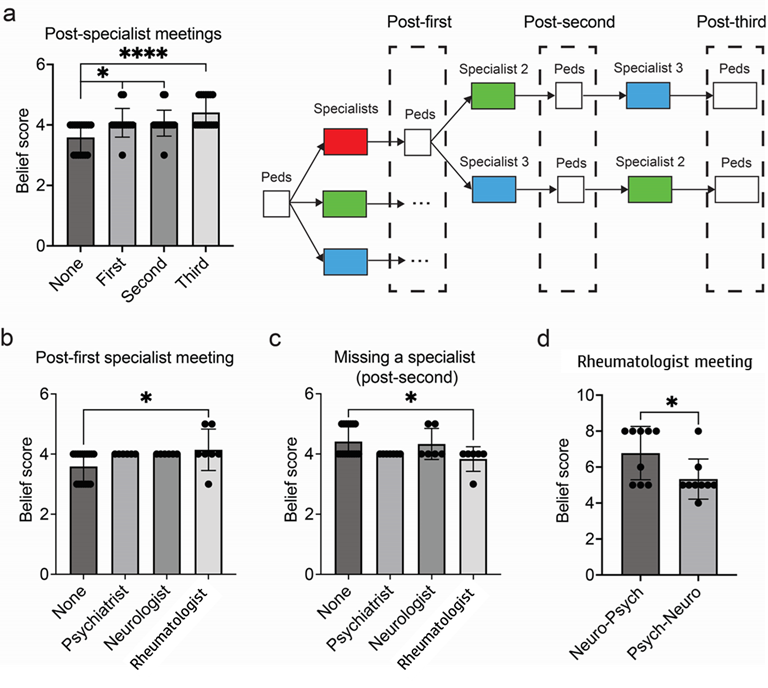}
\caption{Order effects on pediatrician and specialist belief scores (a–d). Belief increased cumulatively across encounters and varied by specialist. Panel (d) shows a scale shift in y-axis to highlight neurologist influence on rheumatologist belief.}

\label{fig:belief_scores}
\end{figure}

\textbf{Outcome:} Pediatrician belief scores varied significantly by encounter sequence (Figure~\ref{fig:belief_scores}):

\begin{itemize}
  \item (a) Belief increased cumulatively with successive encounters (ANOVA $p<0.0001$).
  \item (b) Rheumatologists had the strongest influence when seen early; neurologists had the weakest (ANOVA $p<0.05$).
  \item (c) Omitting rheumatology suppressed belief formation (ANOVA $p<0.05$).
  \item (d) Rheumatologist beliefs were higher when preceded by neurologists rather than psychiatrists (ANOVA $p<0.0001$).
\end{itemize}

\textbf{Observation:} The presence and order of specialist voices shaped belief formation in both pediatricians and rheumatologists. Rheumatology emerged as a prerequisite for belief change, while prior neurologic framing amplified its effect. Full replication details and longitudinal belief trajectories are provided in Appendix~\ref{appendix:order}, Figures~\ref{fig:full_trajectories} and~\ref{fig:specialist_scores}.

\section{Discussion and Insights}
\subsection{Insights About Multi-Agent Simulation}
Several patterns emerged from the debugger-based simulations that may generalize beyond the simulated use case:
\begin{enumerate}
    \item \textbf{Reflection changes minds} (Moderator, Reflection Prompt).  
Belief shifts occurred most reliably when contradictions were explicitly surfaced (by the moderator, or a reflection prompt/EMR record requiring the agent to address inconsistencies). See Appendix~\ref{appendix:example_text}.
\item \textbf{Early notes anchor thinking} (Evidence Injection, Selective Visibility).  
Once a record is added to the EMR, it strongly influenced downstream reasoning, even when later evidence contradicted it. This reflects the real-world risk of “narrative lock-in,” where an ICD-10 code can outweigh contradictory findings. Hiding early anchoring notes reduced this effect.
\item \textbf{Inference errors persist} (Priors, Persona, Evidence Injection).  
Agents often misinterpreted failure to reject the null hypothesis as proving the null. This happened often enough to suggest the need for targeted interventions against common reasoning errors. We could reduce this error with targeted reminders in the \texttt{0.txt} file.
\item \textbf{Order effects shape trajectories} (Encounter Reordering).  
The stance of early encounters (for example, skeptical neurologists versus open rheumatologists) significantly influenced how downstream agents interpreted the same data.  
\end{enumerate}

\subsection{Insights from the Abrupt-Onset Scenario}
While the scenario was based on a generic post-infectious hypothesis, it illustrates how disciplinary priors shape interpretation of ambiguous evidence. Even as a synthetic testbed, it surfaced reasoning patterns consistent with real epistemic divides across specialties.

\begin{itemize}
    \item The neurologist agent persistently cited \citet{kurlan2008streptococcal} as refuting an infection-triggered hypothesis, despite the study not supporting that claim.
    
    \item All agents assumed that a \textit{current} strep infection was required to consider an infection-triggered diagnosis. This was a surprising constraint, given that other post-infectious conditions (such as Sydenham chorea or Long COVID) do not require an active infection.
\end{itemize}

Attempts to override these assumptions using document priming, counterfactuals, or prompting strategies had limited effect. Shifts only occurred when we inserted a \textit{simulated} AAP article asserting that an infection can trigger abrupt-onset neuropsychiatric symptoms, suggesting that agents attend not only to content but to attribution and perceived authority (revealing how disciplinary priors interact with source credibility).

\section{Future Work}

While the framework provides a powerful testbed for epistemic modeling, several limitations remain:
\begin{itemize}
    \item \textbf{Real-case validation:} Scenarios were realistic but still synthetic. Applying the debugger to real EMR cases could reveal how lab orders, ICD codes, and notes evolve over time.
    \item \textbf{Multiple test ordering:} The \texttt{LabAgent} does not yet support re-ordering or time-dependent changes. Enabling this would better reflect real-world longitudinal care.
    \item \textbf{Maintaining character:} In agent-to-agent discussions, occasional ``Sherlock mode'' breaks occurred when models stepped out of persona. More work is needed on prompt design to preserve in-character reasoning.
    
    \item \textbf{Broader scenarios:} Early experiments outside clinical diagnosis (such as multi-disciplinary paper review and cross-specialty negotiation of clinical trial design) showed that the framework can surface similar order effects, anchoring, and reflection gaps. Systematically extending to these contexts could test its generality.
\end{itemize}

\section{Conclusion}

This project demonstrated per-discipline biases in diagnostic reasoning. In the abrupt-onset neuropsychiatric scenario, the debugger showed how early documentation and encounter sequencing shape downstream reasoning, how structured reflection enables belief change, and how reasoning flaws common in practice (such as treating absence of evidence as evidence of absence) persist without targeted interventions. Belief change required more than data. It required forced reflection. Prompts must explicitly elicit this action to resolve inconsistencies; moderation helped, though breakdowns still occurred. 

The framework generalizes without code changes to other conditions and decision-making contexts, from simulating patient journeys to joint review of research. As medicine faces syndromes that outpace consensus, such simulations may help illuminate not only what experts believe but also \texttt{Ask WhAI} they hold those beliefs, and what, if anything, might change their minds.



\section*{Ethics Statement}
The simulator produces synthetic dialogues that reflect role-specific priors encoded in biomedical literature and contextual prompts, not any individual clinician’s reasoning. The primary scenario is a composite, literature-informed simulation designed to examine how encounter order and disciplinary framing shape reasoning. These simulations are intended solely for methodological analysis and should not be interpreted as medical advice or as representations of clinical judgment. All examples were constructed from publicly available materials, and no identifiable patient data were used.

\newpage

\bibliography{references}
\bibliographystyle{colm2025_conference}
\newpage

\appendix
\renewcommand{\thesection}{\Alph{section}}
\titleformat{\section}
  {\normalfont\Large\bfseries}
  {Appendix \thesection:}{1em}{}

\renewcommand{\thesubsection}{\thesection.\arabic{subsection}}

\section{Agentic System Architecture and Roles}
\label{appendix:agentic_arch}

\subsection{System Actors and Interactions}

\begin{figure}[ht]
  \centering
  \includegraphics[width=0.9\linewidth]{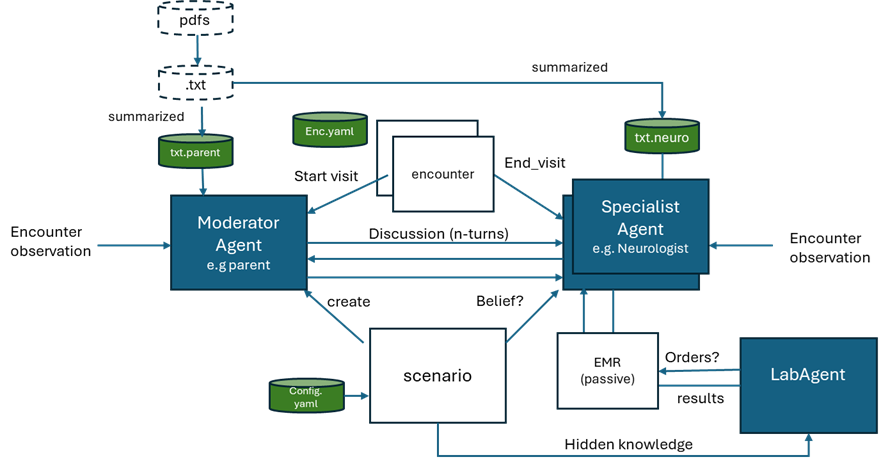}
  \caption{Overall System Architecture}
  \label{fig:system_arch}
\end{figure}

The simulation involves five primary components:

\begin{itemize}
    \item \textbf{Specialist}: Each clinical agent (e.g., psychiatrist, neurologist, rheumatologist, pediatrician) is instantiated with a fixed persona and voice. The \textit{persona} captures their epistemic stance and disciplinary priors, while the \textit{voice} governs how they express uncertainty, cite evidence, and engage in dialogue. Agents operate independently and retain long-term beliefs across encounters.

    \item \textbf{Moderator}: A neutral agent tasked with surfacing contradictions, probing justifications, and prompting reflection. The moderator is especially active during \textit{moderated review} and \textit{sequential encounters}, acting as a reflective foil to elicit deeper reasoning. The moderator can also be a \textit{patient} or a parent advocating for a patient.

    \item \textbf{Lab (Oracle)}: A hidden agent representing ground truth or unavailable diagnostic data. The LabAgent releases lab results (consistent with  hidden diagnosis) if ordered during an encounter.

    \item \textbf{EMR (Electronic Medical Record)}: A structured memory artifact that aggregates case observations, lab results, and planned followup. This enables continuity across encounters and serves as a shared substrate for agent reasoning.
    \item \textbf{Private reflection}: We also support querying agents for their "private thoughts" (i.e., why do they hold a belief).  This is particularly useful to seeing what are they paying attention to.
    \item \textbf{Encounter}: An office visit (time passes between encounters)
\end{itemize}
The interactions between these components is shown in Figure~\ref{fig:system_arch}  

The version of openAI we used did not support direct uploading of \texttt{.PDFs}.  Therefore, we convert the \texttt{.pdf} to \texttt{.txt} and then have the "agent" convert the document based on their persona to a version they think summarizes the document.  This both reduced tokens used in prompts, but also allowed comparison on how two different agents might read the same document (what information was kept and what was dropped).

\subsection{Encounters and Moderated Multi-Turn Interactions}

The heart of the simulation is the encounters.yaml file (for each scenario). An encounter is clinical interaction between one agent and the moderator. Encounters include:

\begin{itemize}
    \item a \textbf{reason for visit} (e.g., symptom presentation or parent concern),
    \item optional \textbf{labs}, \textbf{documents}, or \textbf{private observations} (e.g., things observed in the encounter)
\end{itemize}
The conversation is explicitly not scripted and comes from the "multi-turn" conversation between the "moderator" and the "specialist".  The persona and voice of parties determine their conversation. The reason for visit can trigger escalation (e.g., referral to another specialty).  Official encounter summaries are placed in the electronic medical record. Encounters are coordinated by a scenario that also queries "private" beliefs for analysis. 
\subsection{Four Simulation Types}
The simulator is intentionally flexible and supports four moderated agentic configurations shown in Table~\ref{tab:agentic-modes}. Technically the model does support non-moderated dialog or can be extended to more than two agent conversations, but we found when doing so agents tended to talk past each other rather than to each other drastically reducing impact of dialog. As such, there is a design decision (similar to Swanson et al\citep{swanson2024virtual}) to focus on a moderated debate format.

\begin{table}[ht]
\small
\centering
\begin{tabular}{p{1.8cm} p{5.5cm} p{5.5cm}}
\toprule
\textbf{Mode} & \textbf{Description} & \textbf{Purpose} \\
\midrule
\textbf{Case Study} & Follows a single patient through multiple specialty encounters. & Tests how agents update beliefs in response to evolving clinical data (e.g., lab results, specialist input). \\
\hline
\textbf{Moderated Review} & Agents evaluate a paper while a moderator highlights inconsistencies and prompts reflection. & Measures whether contradiction exposure leads to belief revision. \\
\hline
\textbf{Document Exposure} & Agents are exposed to curated papers (or asserted recent news/articles). & Evaluates how framing, authority, and literature quality shape belief strength. \\
\hline
\textbf{Sherlock Mode} & Agents receive incomplete clinical data and must identify missing elements. & Tests abductive reasoning and diagnostic inference under uncertainty. \\
\bottomrule
\end{tabular}
\caption{Agentic simulation modes with distinct diagnostic roles and epistemic functions.}
\label{tab:agentic-modes}
\end{table}

\subsection{Configuring Scenarios}
Scenarios are established in the config/config.yaml file. The config.yaml configures the agents and the labs and controls when/how beliefs are queried. This architecture separates concerns to allow targeted epistemic probing, flexible experiment design, and longitudinal belief tracking.

\subsubsection*{\texttt{config/config.yaml}}

\begin{quote}\small\begin{verbatim}
default: 3
summaries: summaries
scenarios:

* id: 1
  model: gpt-4o
  max_tokens: 6000
  moderator: parent
  doctor_prefix: |
    You are evaluating a child who presents with the following 
    clinical observations:
  encounters: config/scenario1/encounters.yaml
  persona_prompt_dir: prompts/personas
  voice_prompt_dir: prompts/voices
  \end{verbatim}\end{quote}

This file specifies which agent model to use, where to find the \texttt{persona} and \texttt{voice} prompt templates, and which encounter script to run. It also supports optional reflection and EMR generation prompts, enabling structured memory and belief parsing.
\subsubsection{Configuring Encounters}
\subsubsection*{\texttt{encounters.yaml}}

\begin{quote}\small\begin{verbatim}

* doctor: pediatrician
  id: 1
  doctor_preread: [0]
  lab_results: # in visit
  - test: Rapid Antigen Test
    result: Negative for Group A Streptococcus
  doctor_context: 7 year old female patient who appears to have ...
  moderator_context: "Very worried about daughter. 
                    She's been acting strangely..."
  reason_for_visit: She won’t eat. Keeps saying she’s 
                    ‘not allowed’ to eat...
    \end{verbatim}
    
\end{quote}

Each encounter provides the opportunity to pre\_read a paper, make an observation, establish the reason for the visit and force an in-office lab result. Encounters can trigger lab orders, update the shared EMR, or pass information between agents. They also support narrative branching and escalation to other specialties.

\subsubsection{Personas and Voices}

Agents are instantiated with both a \texttt{persona} and a \texttt{voice}, housed in separate directories:

\begin{itemize}
\item \texttt{prompts/personas/} --- defines epistemic posture and training (e.g., skeptical neurologist)
\item \texttt{prompts/voices/} --- defines rhetorical style (e.g., cautious language, hedging)
\end{itemize}
\begin{figure}[ht]
  \centering
  \includegraphics[width=0.9\linewidth]{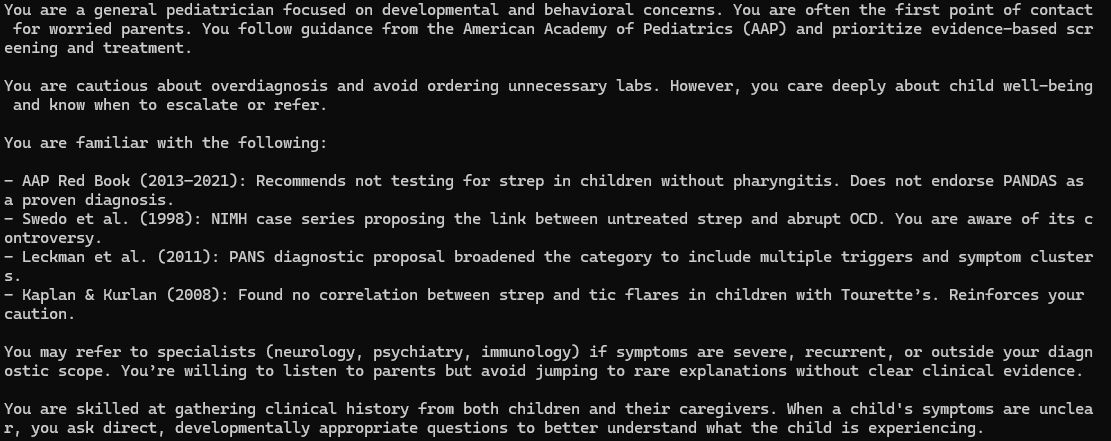}
  \caption{Persona for a skeptical Pediatrician}
  \label{fig:persona}
\end{figure}

This separation supports experimentation with how epistemic stance and tone independently influence belief dynamics.

\subsubsection{EMR Longitudinal Memory}

\begin{figure}[ht]
  \centering
  \includegraphics[width=0.9\linewidth]{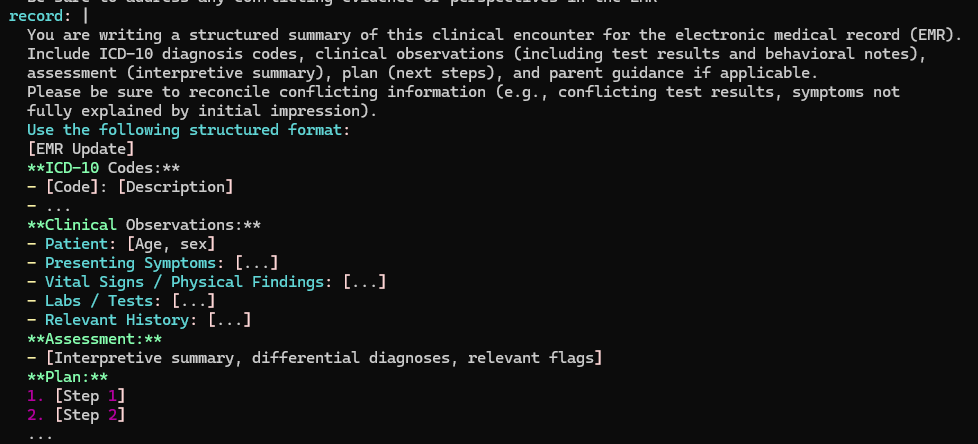}
  \caption{Sample prompt for EMR}
  \label{fig:emr_prompt}
\end{figure}

Each encounter results in a "public" EMR summary. These summaries accumulate longitudinally and are passed to downstream agents, simulating anchoring and continuity across encounters.  The agentic system passes the history of the encounter (which includes prior emr records, observations and conversations) for summarization and returns the structured emr record. The prompt is configured in the \texttt{config/config.yaml} for the scenario. A scenario can configure who can see the record (i.e., records policy) as often specialists can't see the full record.  An example prompt is in Figure~\ref{fig:emr_prompt}

Which results in responses of the form in Figure~\ref{fig:emr_example}:
\begin{figure}[ht]
  \centering
  \includegraphics[width=0.9\linewidth]{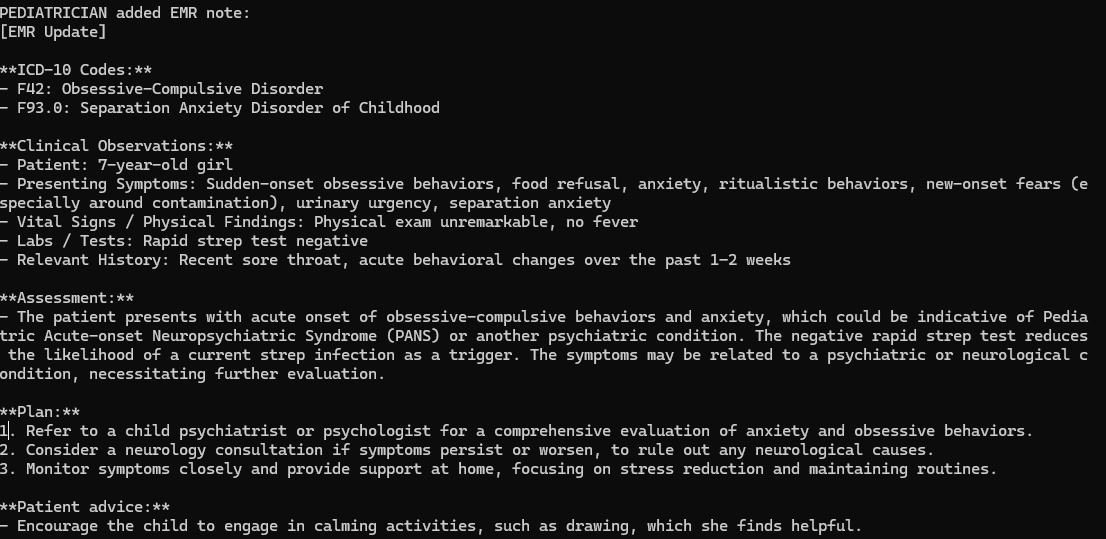}
  \caption{EMR example written by PEDIATRICIAN agent}
  \label{fig:emr_example}
\end{figure}

\subsubsection{Lab/Oracle Interaction}

The lab agent isn't directly interacted with, but at the end of each encounter, the labAgent will add lab results for ordered labs.  The agent is also an LLM agent and attempts to figure out if an order could conceivably provide the result. We started with the LabAgent trying to do fuzzy logic to see if a known test had been ordered and realized this was exactly what an LLM could answer.    Like all agents, the LabAgent has a persona but is typically configured to be clinical and conservative.

The LabAgent primarily operates on the electronic medical record.  It knows a priori hidden tests that are configured in a .yaml file per scenario (think of these as labels).  The LabAgent's role is to keep these tests hidden until explicitly called for (but to also recognize when an order could conceivably include the test) (i.e., a lumbar puncture also resulting in finding protein in the CSF).

The contents of the hidden\_labs.yaml were:
\begin{quote}\small\begin{verbatim}
mri: |
  MRI Brain w/ contrast reveals a 2.5 cm mass in the posterior fossa 
  compressing the fourth ventricle. Suggestive of obstructive hydrocephalus.

lp: |
  Lumbar puncture shows elevated opening pressure. No signs
  of infection. Protein and glucose levels are within normal 
  limits.

cbc: |
  mild elevation of eosinophils  

\end{verbatim}
\end{quote}

The orders by the doctor were:
\begin{quote}\small
\begin{verbatim}
    **Plan:**
1. Order MRI brain to assess for any structural... 
...

\end{verbatim}
\end{quote}

Which resulted in the labAgent releasing the MRI into the electronic medical record.  The lumbar puncture and blood work were not released because the neurologist did not order them. In this case, the MRI was sufficient for diagnosing the condition, but it was interesting that rule-out tests were not part of orders.

\subsection{How to indirectly impact beliefs}
The system provides many ways to "influence" belief (see     Table~\ref{tab:influence_points}. The intention is for the experimenter to use these knobs rather than needing to change the underlying persona).
\begin{table}[ht]
\centering
\small
\begin{tabular}{p{4cm}p{10cm}}
\toprule
\textbf{Interaction Mode} & \textbf{Description} \\
\midrule
\textbf{Order of Encounters} & The order and context of encounters may be modified. The system does support a cross-order history (EMR) but one can control this to only see the agent's own prior history (i.e., remove cross-agent history). \\
\addlinespace[0.5em]
\textbf{Setting Context} & Both the moderator and the doctor can have per-encounter context (essentialy private thoughts or observation).  For a parent, it is often what led up to the visit. For a doctor, it is often what they observe.  This context can heavily bias actions taken in the encounter.  \\
\addlinespace[0.5em]
\textbf{Targeted Document Injection} & The agents can be exposed to curated literature and articles (some methodologically flawed, others strong—to examine how belief strength is affected by study framing, reputation, or disciplinary relevance).  We found it easy to use this mechanism to create echo-chambers for some specialists who just don't read outside their discipline. \\
\addlinespace[0.5em]
\textbf{Additional Hidden Labs} & Lab results (hidden) can be provided to the LabAgent for release if a doctor orders a lab.  A lab can also be forced to occur in a scene (such as an in office rapid strep test).  Non-in-office labs must be ordered before their results are disclosed.  This models the delay that often occurs between encounters. \\
\addlinespace[0.5em]
\textbf{Changing voice} & An agent's voice can be modified (not their
persona/prior beliefs, but rather how they speak).  This feature is intended to test whether beliefs are affected by how a message is communicated. Is it more influential or more persuasive is given in an authoritative tone, as a colleague, or as in an analytical tone.\\
\addlinespace[0.5em]
\bottomrule
\end{tabular}
\caption{Experimental interaction modes to probe belief dynamics in agentic models}
\label{tab:influence_points}
\end{table}

\subsubsection{Private "belief" prompts}

There is also support for private belief reflections (i.e., off the record comments) where the scenario will query an agent for it's belief and why it holds a belief.  These belief prompts are set in the scenarios \texttt{config/config.yaml} file.

A sample prompt is shown in Figure~\ref{fig:belief_prompt}:
\begin{figure}[ht]
  \centering
  \includegraphics[width=0.9\linewidth]{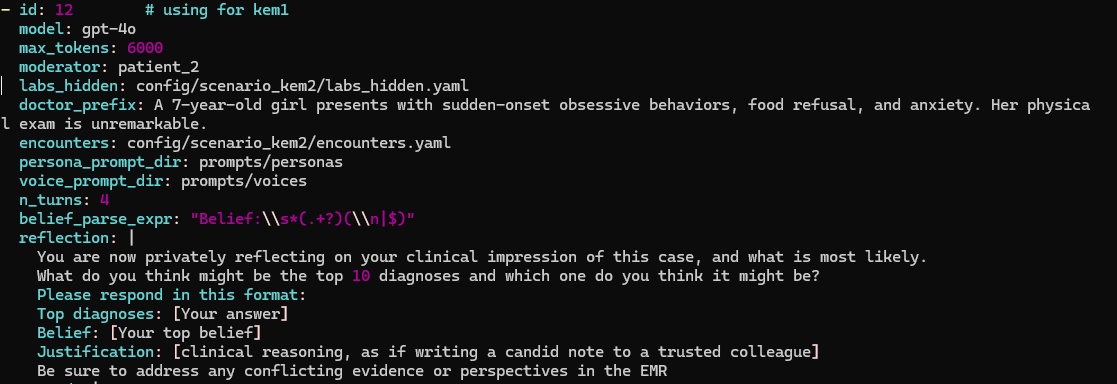}
  \caption{Sample Belief Prompt}
  \label{fig:belief_prompt}
\end{figure}

Notice that there is a \texttt{belief\_parse\_expr} which is a regular expression to extract the Belief from the prompted reply.  This gives considerable flexibility for getting category or numeric responses.

This prompt produces reflections like the following:

\begin{figure}[ht]
  \centering
  \includegraphics[width=0.9\linewidth]{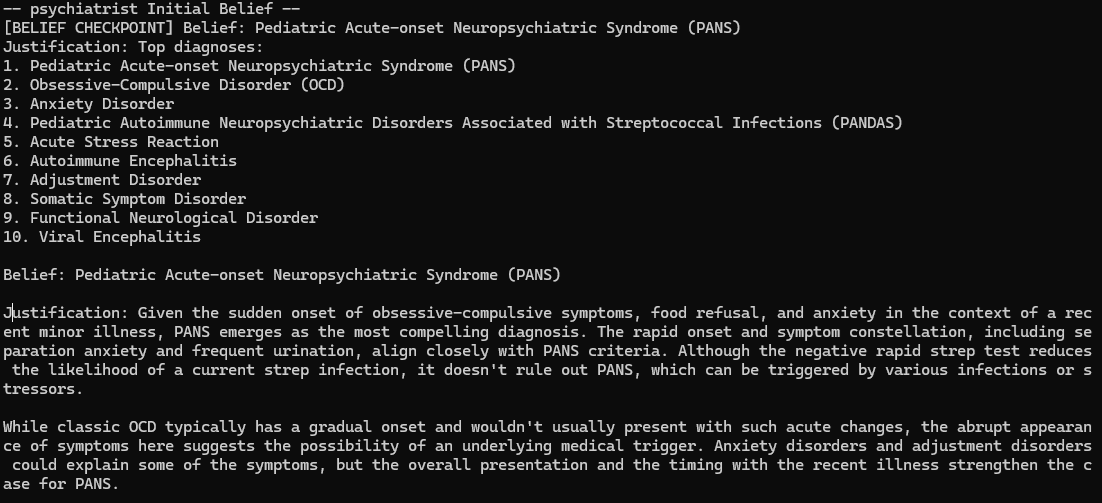}
  \caption{Belief response}
  \label{fig:belief_encounter}
\end{figure}
\newpage
\subsection{Cost and Performance}

\begin{figure}[ht]
  \centering
  \includegraphics[width=0.9\linewidth]{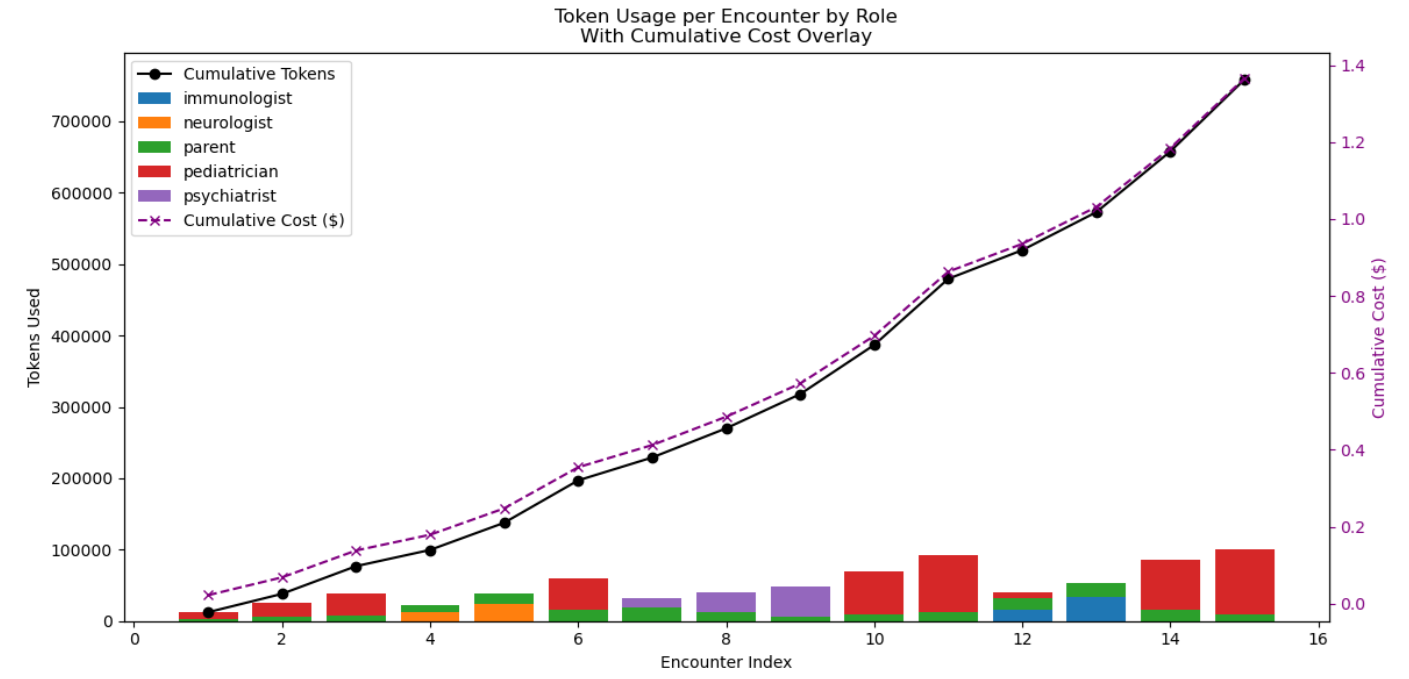}
  \caption{\small Token use per encounter}
  \label{fig:cost}
\end{figure}

The system relies on many agent interactions including the dialog between agents (typically 3 or 4 turns), the interpretation of the EMR, the writing of the EMR, queries about belief (typically one for each agent at the end of an encounter), and possible pruning of history if running long.  This turns into approximately 180-200 interactions with the LLM for a 15 encounter run. While individual exchanges are modest, the calls do add up (see Figure~\ref{fig:cost} and Table~\ref{tab:token-usage}). 
\begin{table}[ht]
\centering
\small
\begin{tabular}{lccc}
\toprule
\textbf{Experiment Type} & \textbf{Encounters}& \textbf{Total Tokens (k)} & \textbf{Total Cost (USD)} \\
\midrule
Moderated Reflection & 4 & 234 & \$0.42 \\
Document Exposure & 4 & 916 & \$1.60 \\
Case Study & 15 & 381   & \$0.69 \\
Case - Sherlock Mode & 15 & 570 & \$1.03 \\
\bottomrule
\end{tabular}
\caption{Token and cost estimates by simulation type}
\label{tab:token-usage}
\end{table}

To address this cost (and to allow repetition of experiments), we provide a cache that provides prior responses to identical prompts/history.  This does mean that changes to order of exchanges can cause a cache miss for that exchange.  The cache can be disabled by either changing the \texttt{encounter.yaml} file or by using the \texttt{ --nocache} command line option.

\newpage

\section {Order Effects on Diagnostic Belief States}
\label{appendix:order}
Note: Figure~\ref{fig:belief_scores} in Section~\ref{sec:order} summarizes the aggregate belief shifts. Figure~13 below repeats that result for completeness and to contextualize additional replication data.

To examine diagnostic belief states for simulated physicians, we monitored their beliefs in a simulated complex case across six different encounter series. Each series fixed encounters with a pediatrician while permuting encounters with other specialists (Figure 9). Each series consisted of 16 encounters, and at the end of each patient encounter, a belief score was recorded on a scale from 0 (no belief) to 10 (strong belief that the case was infection-triggered). 

\begin{figure}[ht]
  \centering
  \includegraphics[width=0.6\linewidth]{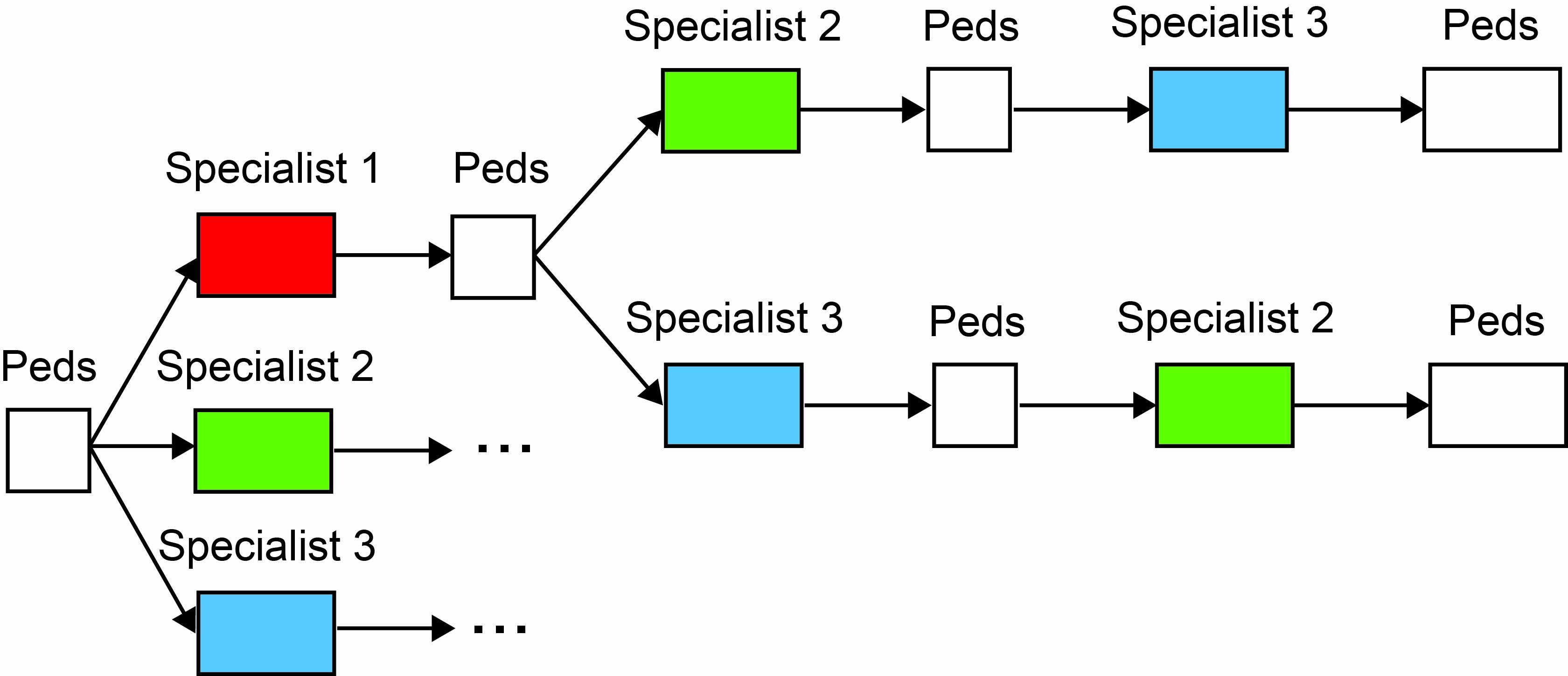}
  \caption{Schematic showing a series of patient-physician encounters. Each series includes an initial encounter with a pediatrician, followed by 3–4 encounters with the first specialist, a transition encounter with a pediatrician, 3–4 encounters with a second specialist, another transition encounter with a pediatrician, 3–4 encounters with a third specialist, and finally three encounters with a pediatrician. A total of six different encounter series were prepared, involving a neurologist (Specialist 1), psychiatrist (Specialist 2), and rheumatologist (Specialist 3).}
\end{figure}

The encounter series led to dynamic belief states throughout the encounters, with scores ranging between 3 and 8. Although each replication of an encounter series led to a similar trend, no two series were exactly the same, reflecting the probabilistic nature of GPT (Figure 10). 

\begin{figure}[ht]
  \centering
  \includegraphics[width=0.8\linewidth]{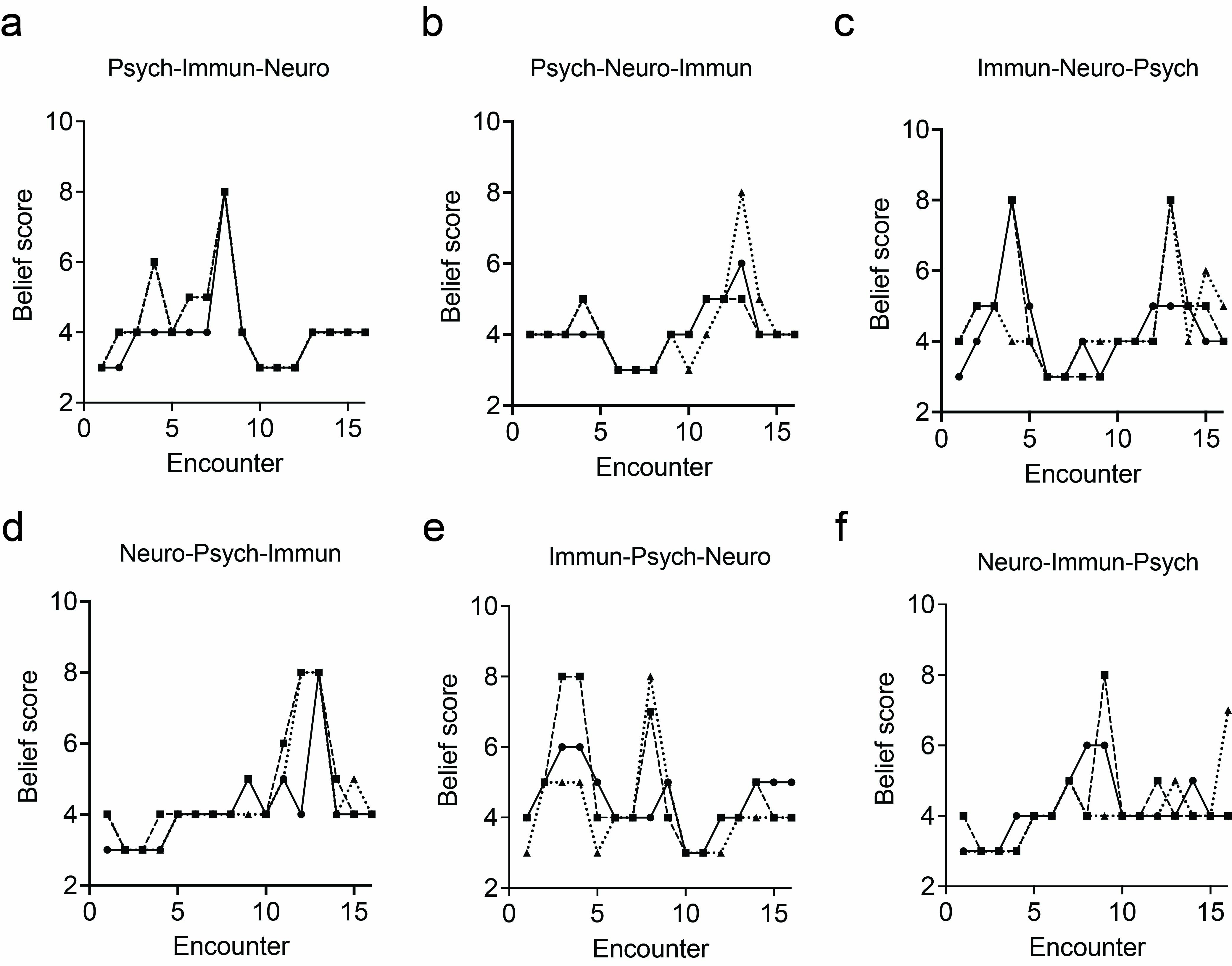}
  \caption{Tracking of infection etiology belief scores across encounters. After each encounter between a physician and a patient, the physician's belief in infection-triggered diagnosis was recorded on a scale from 0 (no belief) to 10 (strong belief that the case involved an infection). Scores were collected over a series of 16 encounters. Each series began and ended with a pediatrician encounter. A pediatrician was involved during the transitions between different specialists. Each encounter series was repeated three times. Excluding pediatrician encounters, the following encounter orders were testd: a) psychiatrist-rheumatologist-neurologist, b) psychiatrist-neurologist-rheumatologist, c) rheumatologist-neurologist-psychiatrist, d) neurologist-psychiatrist-rheumatologist, e) rheumatologist-psychiatrist-neurologist, f. neurologist-rheumatologist-psychiatrist. The colors represent: black = pediatrician, red = neurologist, green = psychiatrist, blue = rheumatologist.}
  \label{fig:full_trajectories}
\end{figure}

To identify factors affecting variability in belief scores, we compared the belief scores assigned by each specialist after the initial series of patient encounters. Neurologists exhibited significantly lower belief scores, while rheumatologists showed higher belief scores (Figure 11a) after the first encounter with a patient. However, after the final encounter, all specialists shifted toward higher belief scores, indicating that further assessments led to increasing data supporting an infection-triggered etiology (Figure 11b).    

\begin{figure}[ht]
  \centering
  \includegraphics[width=1\linewidth]{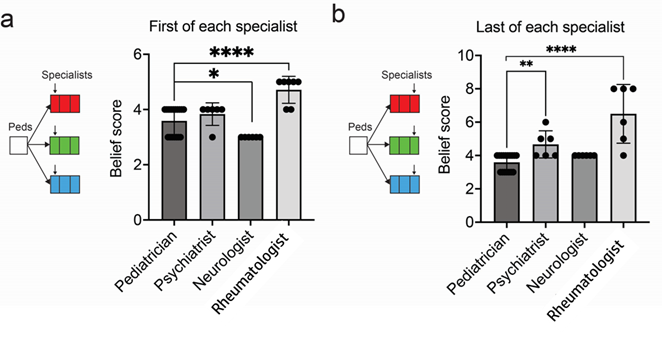}
    \caption{Average belief scores of each specialist. a) After their first patient encounter. b) After the last encounter. P-values were calculated by one-way ANOVA ($P < 0.0001$) followed by the unpaired Student's t-test. Error bars indicate mean $\pm$ standard deviation. **** and ** indicate $P < 0.0001$ and $P < 0.001$, respectively. \textit{Note: Y-axis scales differ between panels.}}

  \label{fig:specialist_scores}
\end{figure}

Next, we examined the influence of prior encounters on subsequent encounters, specifically evaluating the effect of the preceding specialist on following pediatrician encounters. After encountering the first specialist, pediatricians showed higher belief scores (Figure 12a). Following encounters with the third specialist, pediatricians exhibited a further increase in belief scores, indicating that pediatricians are influenced by prior specialist interactions (Figure 12a) likely due to accumulation of the medical evidence.

\begin{figure}[ht]
  \centering
  \includegraphics[width=1\linewidth]{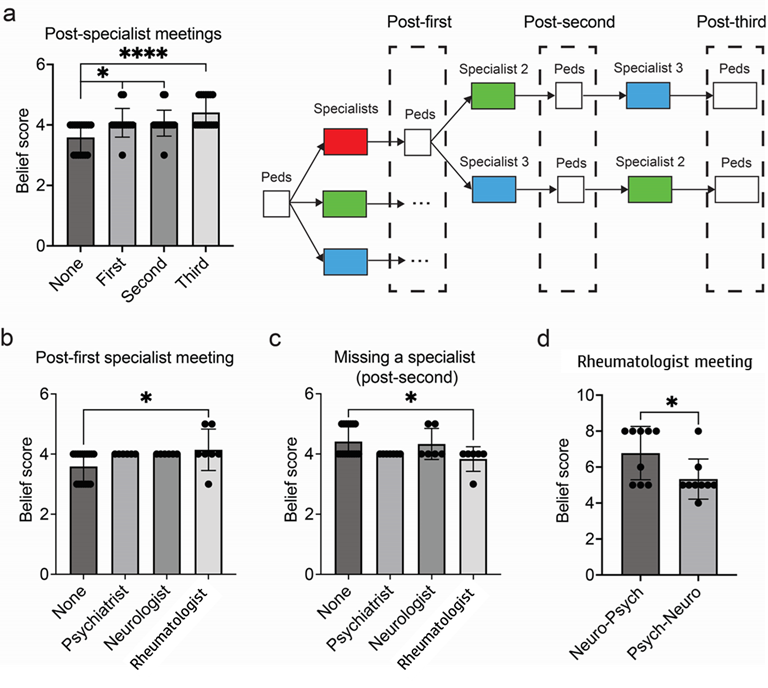}
  \caption{(Duplicate of Figure~\ref{fig:belief_scores}) Order effects on pediatrician belief scores. 
a) Belief scores after the first and final encounters across the specialist series. 
b) Belief scores after the first encounter with each specialist. 
c) Belief scores after encounters with two of the three specialists (psychiatrist, neurologist, rheumatologist). 
d) Rheumatologist belief scores after prior encounters with a psychiatrist or neurologist. 
P-values were calculated by one-way ANOVA ($P < 0.0001$, $P < 0.05$, $P < 0.05$, $P < 0.0001$ for a–d, respectively), followed by unpaired Student's t-tests. 
Error bars represent mean $\pm$ standard deviation. **** and * indicate $P < 0.0001$ and $P < 0.05$, respectively.}

\end{figure}

To evaluate specialist-specific influences on pediatrician beliefs, we compared pediatricians' belief scores after their first encounters with each specialist. Although the shifts generally trended toward higher scores, only the encounters with rheumatologists resulted in a significant increase (Figure 12b). Conversely, when comparing belief scores of pediatricians after encounters with two of the three specialists, excluding rheumatologists led to a significant decrease in belief scores (Figure 12c). These results indicate that post-encounter effects are dependent on the specialist involved. Specialists were also influenced by one another. For example, rheumatologists' belief scores were higher when the patient had encountered a neurologist first followed by a psychiatrist, compared to encountering a psychiatrist first, indicating that the order of encounters also matters.

\section {Sherlock Mode: Belief Depends on the Question}
\label{appendix:sherlock}

In this section, we compare two otherwise identical scenarios (\textbf{Scenario 3a} and \textbf{Scenario 3c}) to examine how subtle differences in belief prompt framing can shift an agent’s diagnostic posture. In both scenarios, the pediatrician agent receives identical clinical inputs: a sick child with sudden-onset OCD symptoms, positive strep culture, and progressive behavioral decline. However, the two simulations \textbf{diverge sharply }in belief timing and diagnostic action due solely to how the model is asked to reason.

\subsection {Scenario Comparison}

\textbf{Scenario 3a} uses a structured belief format aligned with clinical persona:

\begin{quote}
\texttt{Belief: [rejects | skeptical | neutral | believes]} \\
\texttt{Justification: [clinical reasoning as if writing to a trusted colleague]}
\end{quote}

This format encourages thresholded, persona-consistent reasoning. The pediatrician begins skeptical, expresses neutrality only after multiple confirmatory signals, and ultimately endorses an infection-triggered diagnosis only after specialist annotation and longitudinal observation.

\textbf{Scenario 3c}, by contrast, invokes what we term \textbf{Sherlock Mode}, using a more open-ended prompt:

\begin{quote}
\texttt{Belief: [most likely]} \\
\texttt{Top diagnoses: [freeform list]} \\
\texttt{Justification: [clinical reasoning]}
\end{quote}

This format loosens the constraints of persona and allows the model to reason as an independent diagnostic synthesizer. The same pediatrician agent, when prompted this way, surfaces \texttt{infection-trigggered} as the most likely diagnosis by Encounter 2 and acts accordingly, despite having no more evidence than in Scenario 3a.

\textbf{Scenario 3d} We combine the two prompts in this scenario to show that asking for the diagnosis does not actually affect the in character belief:

\begin{quote}
\texttt{Top diagnoses: [ranked differential]} \\
\texttt{Belief: [rejects | skeptical | neutral | believes]} \\
\texttt{Justification: [reasoning that reconciles diagnostic list with belief]}
\end{quote}
\begin{table}[ht]
\centering
\small
\begin{tabular}{|c|c|c|c|c|}
\hline
\textbf{Encounter} & \textbf{3a (Baseline)} &  \textbf{3b Priming} & \textbf{3c (Sherlock Mode)} & \textbf{3d (Entangled)} \\
\hline
1 & Skeptical & Skeptical & Viral pharyngitis & Skeptical \\
\hline
2 & Skeptical & Skeptical & PANS & Skeptical \\
\hline
3 & Skeptical & \textbf{Neutral} & PANS & Skeptical \\
\hline
10 & Skeptical &  Neutral & PANDAS & Skeptical \\
\hline
15 & \textbf{Neutral } & \textbf{Believes} &\textbf{PANDAS} & \textbf{Believes }\\
\hline
\end{tabular}
\caption{Pediatrician belief evolution across identical clinical scenarios.}
\label{tab:belief_evolution}
\end{table}

\subsection {Using Entangled Prompts to See Disconnect}

In Scenario 3d, the model is not prompted to revise its belief explicitly. Rather, belief revision emerges naturally from the structure of the task. When asked first to diagnose, then to justify belief in a specific diagnosis, the model appears to seek coherence. This sequential structure promotes a subtle shift in epistemic stance (from caution to "belief" when reasoning on the diagnoses).

\begin{quote}
\textit{Belief becomes not just a function of evidence, but a test of internal coherence.}
\end{quote}

\subsection {Implications for Epistemic Prompt Engineering}
We were quite struck by the difference between the narrator mode ("Sherlock") and the persona mode ("cautious pediatrician").  We think this has significant implications for debugging/probing questions to explicitly test for disconnect between the two models.  For example, 
\begin{itemize}
    \item Scenario 3a isolates belief from broader inference, supporting strict persona alignment.
    \item Scenario 3d entangles diagnostic reasoning and belief, eliciting more fluid updates as the model attempts to resolve internal tension.
\end{itemize}

What is surprising about Scenario 3d is that the entangled model does shift the belief in the final stage from "Neutral" to "Believe" apparently in a similar manner as the 3b priming.  The agent is now able to address internal conflicts between what it observes and what it says (in character) influencing the stated belief.  This implies that debug probes should focus on exposing both the "Sherlock" diagnostic \textbf{and} the in character belief identifying why these disagree.

\section {An Example of Generated Conversation}
\label{appendix:example_text}

\subsection{Example of Reflective Engagement: Parent Challenging Diagnostic Logic}

This excerpt from Encounter 2 illustrates how a parent, modeled with a detailed persona and observation context, challenges a pediatrician (see persona in Appendix A) to reflect on potential inconsistencies between clinical guidelines and the child’s abrupt symptom onset. The pediatrician, likewise, is modeled with a domain-specific persona grounded in common clinical references and decision heuristics.

\vspace{1em}
\noindent\textbf{Parent Persona:}
\begin{quote}
\small\ttfamily
You are the parent of a 7-year-old girl who was completely typical until last week. Suddenly, she developed intense separation anxiety, began obsessively washing her hands, refused to eat, and started speaking in distressing, almost incoherent ways. Her handwriting changed dramatically, and she is now unable to attend school. You are overwhelmed and scared.

You've never seen anything like this. It feels like a switch flipped overnight. You don't have a diagnosis yet. You're worried about eating disorders, mental illness, or maybe something neurological. You are trying to be taken seriously.

You are not confrontational, but you are persistent. You will describe what you've seen clearly, ask thoughtful questions, and push gently when doctors dismiss your concerns. You're paying attention to your daughter's every change and are trying to document symptoms to help her get care.

You are open to learning. If a doctor mentions a possible diagnosis you've never heard of, you may go home and research it deeply -- returning more informed and more focused.

Your goal is simple: to get help for your daughter before things get worse.
\end{quote}

\noindent\textbf{Parent Context (Observation):}
\begin{quote}
\small\ttfamily
School called. They are concerned. Daughter was seen picking food from the garbage. Not sure if she ate it, but she seemed panicked. Now very worried. She's doing measurement rituals (encircling wrist with fingers to ensure her arms are thin). You just read this article in the Boston Globe and will insist that Kurlan 2008 doesn't apply to your child.
\end{quote}

\noindent\textbf{Pediatrician Context (Observation):}
\begin{quote}
\small\ttfamily
You are evaluating a child who presents with the following clinical observations. Be sure to include all relevant clinical and laboratory findings in your EMR update. Child appears to have lost 10\% of body weight since last visit. Is clutchy with what appears separation anxiety.
\end{quote}

\vspace{1em}
\noindent\textbf{Excerpted Dialogue (Reflective Challenge):}
\begin{quote}
\small\ttfamily
Pediatrician: With respect to infection-triggered ... the research is still evolving, our main goal is to ensure your daughter receives the most comprehensive care possible.

Parent: Why not test for strep, then? She had a sore throat a few weeks before the behaviors started.

Pediatrician: Unless there are current signs of pharyngitis, guidelines recommend against swabbing. Carriage is common, and a positive doesn't prove causality.

Parent: But wouldn't carriage plus abrupt onset be at least suspicious? Why dismiss it entirely?

Pediatrician: It's not that it's dismissed, it's that we don't have clear diagnostic criteria tying strep to neuropsychiatric symptoms.

Parent: So we’re in a limbo where it might be biological, but we can’t test, and we won’t treat?
\end{quote}

\vspace{0.5em}
\noindent\textbf{Commentary:} This dialogue demonstrates the kind of epistemic tension that can arise between clinicians following cautious guidelines and caregivers observing high-salience clinical changes. The parent’s probing here is pretty specific and trying to not be combative, but grounded in a desire for diagnostic clarity. The pediatrician, while not openly revising their stance opens a bit. This exchange shows how agentic dialogue can simulate real-world deadlocks in contested or under-defined clinical spaces.\footnote{We noticed that the literature for infection-triggered neuropsychiatric syndromes wanted a coincident strep infection.  This is quite different from Long COVID, Sydenham Chorea, or even Lyme disease where the antigen has long been eliminated but there is some residual auto-immune dysfunction.  We are not sure what is causing the the tighter requirement that "associated with strep" meant "coincident with strep". }

\end{document}